\documentclass[letterpaper, 10 pt, conference]{ieeeconf}  % Comment this line out if you need a4paper

\IEEEoverridecommandlockouts                              % This command is only needed if 
\usepackage{float}
\usepackage{graphicx}
\usepackage{amsmath}
\usepackage{amssymb}
\usepackage{tabularx}
\usepackage{bm}
\usepackage{lipsum}
\usepackage{mathtools, siunitx}

\newcommand{\vect}[1]{\boldsymbol{\mathbf{#1}}}

\newcommand{\inno}{\delta\mathbf{z}_k^-}
\newcommand{\eye}[1]{\vect{I}_{#1}}

\newcommand{\e}[1]{\times 10^{#1}}

\DeclareMathOperator{\diag}{diag}

\title{\LARGE \bf
Unscented Kalman Filtering on Manifolds for AUV Navigation - Experimental Results
}

\author{Stephen T. Krauss$^{1}$ and Daniel J. Stilwell$^{1}$% <-this % stops a space
\thanks{$^{1}$Bradley Department of Electrical and Computer Engineering,
        Virginia Polytechnic Institute and State University, Blacksburg, VA, USA
        {\tt\small \{stkrauss, stilwell\}@vt.edu}}%
}

\begin{document}

\maketitle
\thispagestyle{empty}
\pagestyle{empty}

%===============================================================================
% Abstract
%===============================================================================
\begin{abstract}

In this work, we present an aided inertial navigation system for an autonomous underwater vehicle (AUV) using an unscented Kalman filter on manifolds (UKF-M). The inertial navigation estimate is aided by a Doppler velocity log (DVL), depth sensor, acoustic range and, while on the surface, GPS. The sensor model for each navigation sensor on the AUV is explicitly described, including compensation for lever arm offsets between the IMU and each sensor. Additionally, an outlier rejection step is proposed to reject measurement outliers that would degrade navigation performance. The UKF-M for AUV navigation is implemented for real-time navigation on the Virginia Tech 690 AUV and validated in the field. Finally, by post-processing the navigation sensor data, we show experimentally that the UKF-M is able to converge to the correct heading in the presence of arbitrarily large initial heading error. 

\end{abstract}

%===============================================================================
% Introduction
%===============================================================================
\section{Introduction}\label{sec.introduction}

For an autonomous underwater vehicle (AUV) without access to GPS signals while under water, inertial navigation is one of the main sources of localization information. The basic premise behind inertial navigation is that a vehicle equipped with an inertial measurement unit (IMU) can integrate measurements of angular velocity and linear acceleration into estimates of attitude, velocity, and position. However, it is well known that integration of small sensor errors over time causes a buildup of error in the inertial navigation estimates \cite{titterton1997strapdown}. In order to correct the buildup of error, a Kalman filter is commonly used to combine IMU measurements with measurements from other sensors on an AUV, such as a Doppler velocity log (DVL) and depth sensor. While the extended Kalman filter (EKF) is one of the most commonly used filters, other works have demonstrated advantages to using other variants of the Kalman filter, such as the unscented Kalman filter (UKF).

\begin{figure}[!h]
\centering 
\includegraphics[width=\linewidth]{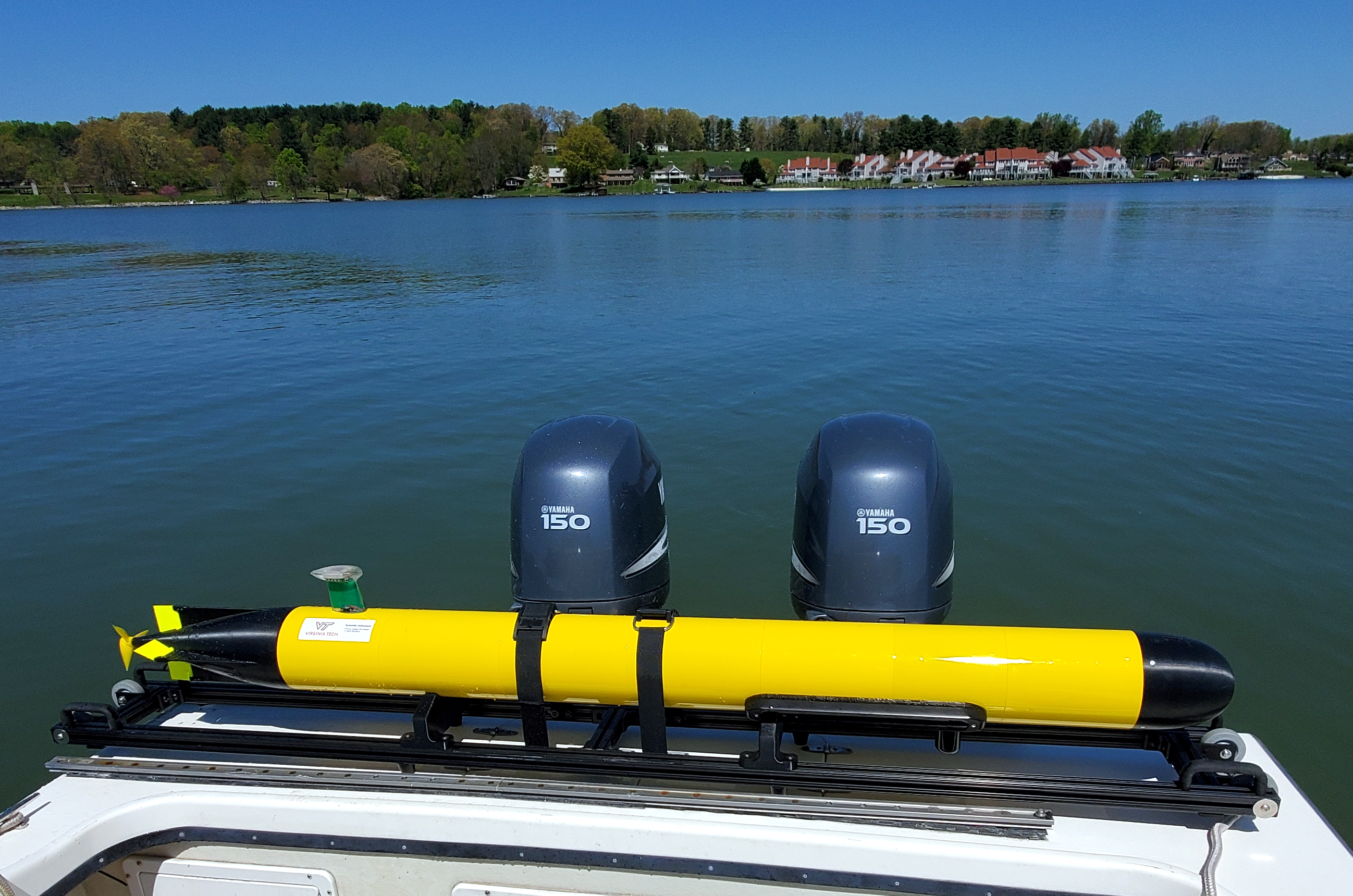}
\caption{Virginia Tech 690 AUV aboard the Virginia Tech research vessel in Claytor Lake, Dublin, VA, USA.}
\label{fig.690_auv}
\end{figure}

In this work, we expand upon the work done in \cite{cantelobre2020real}, where the authors propose the use of an unscented Kalman filter on manifolds (UKF-M) for AUV navigation. First, we explicitly describe the sensor models used in the measurement update step of the UKF-M and include compensation for lever arm effects caused by the mounting locations of the sensors on the AUV. Next, we propose UKF-M retraction and inverse retraction functions that allow position standard deviation to be reported in meters instead of radians. Additionally, we introduce a measurement validation algorithm to the UKF-M measurement update step in order to reject measurement outliers that would otherwise affect navigation accuracy. Finally, we validate a real-time implementation of the filter in the field using the Virginia Tech 690 AUV, shown in Figure \ref{fig.690_auv}, with a mission consisting of surface calibration maneuvers and a survey pattern. Post processing of the dataset is used to demonstrate the ability of the UKF-M to converge in the presence of large initial heading errors. Convergence despite large initial heading errors allows the filter to operate without the need for a separate filter for determining initial conditions.

The remainder of this work is organized as follows. In Section \ref{sec.related_work}, previous works related to this work are summarized. In Section \ref{sec.inertial_navigation_equations}, we introduce the coordinate frames and inertial navigation equations used in this work. In Section \ref{sec.sensor_models}, we present the sensor models for the navigation sensors on the 690 AUV. In Section \ref{sec.manifold_ukf}, a summary of the unscented Kalman filter on manifolds is given, and in Section \ref{sec.outlier_rejection}, a simple measurement outlier rejection algorithm for the filter's measurement update step is presented. Finally, in Section \ref{sec.experimental_results}, results from a real-time implementation of the filter tested in the field with the Virginia Tech 690 AUV are given.

%===============================================================================
% Related Work
%===============================================================================
\section{Related Work}\label{sec.related_work}

One of the most common approaches to reducing or bounding the buildup of inertial navigation error is to incorporate measurements from other sensors through an extended Kalman filter (EKF), such as those presented in \cite{titterton1997strapdown, farrell1998global, groves2013principles} and \cite{miller2010autonomous}. One of the disadvantages of the EKF is that it requires linearization of the inertial navigation dynamics, which reduces accuracy of the estimate of the state and its covariance. Additionally, due to the linearization of the inertial navigation equations, the EKF can be susceptible to divergence if the initial state estimate is not close to the true initial state. 

The most common way this problem arises is through initialization of the heading angle. In order to provide an accurate initial heading angle to the EKF, a two-step approach consisting of coarse alignment followed by fine alignment is commonly used \cite{groves2013principles}. Other approaches to the handling of large heading error with an EKF include \cite{xiaoying1999development}, where an EKF model that can handle large initial alignment errors is developed. This approach still uses an EKF, which requires linearization, and must also switch to a different inertial navigation model when alignment error is reduced below a certain level, much like the coarse and fine alignment approach. In this work, use of the UKF-M allows for filter convergence in the presence of large initial heading error without any special modifications to the filter or system model.

Other variations of the Kalman filter have been proposed in the literature for nonlinear filtering, such as the unscented Kalman filter (UKF) \cite{wan2000unscented}. Instead of linearizing the state propagation and measurement equations, the UKF uses a set of points, named sigma points, distributed such that their empirical mean and covariance estimate the state mean and covariance. These sigma points are propagated through the nonlinear equations in the prediction and update steps. In this way, the UKF can be accurate to a higher order than the EKF and does not require linearization of the inertial navigation equations. Approaches to using the UKF for aided inertial navigation system have been explored in works such as \cite{shin2004unscented}, \cite{barisic2012sigma}, and \cite{allotta2016new}. However, implementation challenges exist when using the UKF for inertial navigation. If vehicle roll, pitch, and heading are represented in the state vector as Euler angles, the algorithm must compensate for the fact that 0 degrees and 360 degrees are equivalent. When propagating sigma points through the inertial navigation equations, one or more of the sigma points may wrap, leading to an inaccurate representation of the state mean and covariance.

The unscented Kalman filter on manifolds (UKF-M), proposed in \cite{brossard2020code}, is a modification of the UKF that is able to directly handle states that evolve on the manifold $SO(3)$, such as vehicle attitude. With vehicle attitude represented by an rotation matrix in $SO(3)$, the effect of wrapping of Euler angles on sigma points can be avoided. In \cite{cantelobre2020real}, the authors present an inertial navigation algorithm for AUVs based on the UKF-M. The authors demonstrate the ability of the filter to accurately estimate the state covariance through Monte-Carlo simulations and show that the filter can run in real time on an AUV. In this work, we build upon the results in \cite{cantelobre2020real} by using the UKF-M for inertial navigation using sensor models with lever arm compensation and validating a the filter in real time with data from an AUV survey mission.

%===============================================================================
% Intertial Navigation Equations
%===============================================================================
\section{Inertial Navigation Equations}\label{sec.inertial_navigation_equations}

We adopt the notation and navigation frame inertial navigation formulations presented in \cite{groves2013principles}. The inertial navigation equations involve three reference frames: the Earth-centered Earth-fixed (ECEF) frame, the body frame, and the navigation frame.

The origin of the Earth-centered Earth-fixed (ECEF) reference frame is attached to the geometric center of the Earth.  The positive $z$-axis of the ECEF reference frame is aligned with the Earth's axis of rotation and passes through the north pole. The positive $x$-axis passes through the intersection of the equator and the prime-meridian. The $y$-axis completes a right-handed coordinate system \cite{titterton1997strapdown}. Note that, since the ECEF frame is fixed to the earth, it is rotating with respect to an inertial frame \cite{farrell1998global}. Values expressed in this frame are indicated with an $e$.

The origin of the body frame is fixed to a point on the AUV with the positive $x$-axis passing through the nose of the AUV, the positive $z$-axis pointing down, and the $y$-axis completing a right-handed coordinate system. In this work, the body frame is taken to be the IMU measurement frame. Values expressed in the body frame are denoted with $b$.

The navigation frame is a coordinate frame that is tangent to the earth's surface with an origin co-located with the origin of the body frame. The navigation frame's north direction points toward true north, the down direction points toward the earth's center perpendicular to the tangent plane, and the east direction points east orthogonal to the north and down directions. Values in the navigation frame are notated with an $n$.

\begin{figure}[ht]
\centering
\includegraphics[width=2.5in]{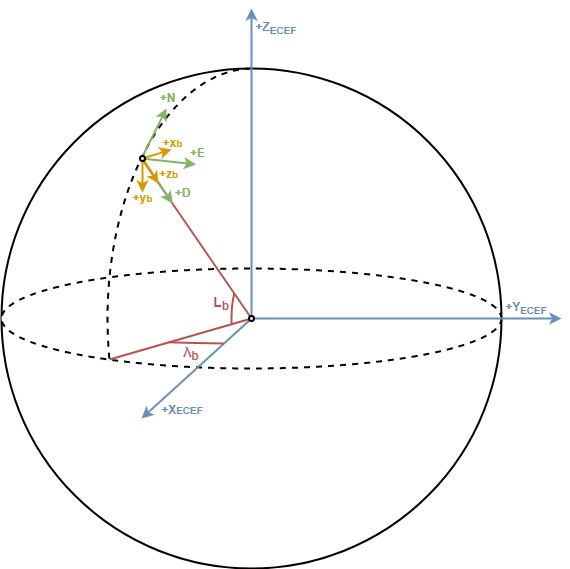}
\caption{Illustration of the inertial navigation reference frames at a moment in time. The navigation frame origin is fixed to the body frame origin with position described by its geodetic latitude ($L_b$) and longitude ($\lambda_b$).}
\label{fig.reference_frames}
\end{figure}	

The nonlinear kinematic equations that relate attitude, velocity, and position of the body frame to angular velocity and linear acceleration of the body frame are
\begin{equation}
\begin{split}
\dot{\vect{C}}_b^n      &= \vect{C}_b^n \vect{\Omega}_{ib}^b - (\vect{\Omega}_{ie}^n + \vect{\Omega}_{en}^n)\vect{C}_b^n \\
\dot{\vect{v}}_{eb}^n   &= \vect{f}_{ib}^n + \vect{g}_b^n - (\vect{\Omega}_{en}^n + 2\vect{\Omega}_{ie}^n) \vect{v}_{eb}^n \\
\dot{\vect{p}}_b        &= \vect{T}_{r(n)}^p \vect{v}_{eb}^n \\
\end{split}
\label{eq.nav_equations}
\end{equation}
where $\vect{C}_b^n$ is the rotation matrix from the body frame to the navigation frame, $\vect{v}_{eb}^n$ is the velocity of the body frame relative to the ECEF frame, expressed in the navigation frame, and $\vect{p}_b$ is the curvilinear position of the body frame. The navigation frame velocity and position are expressed
\begin{equation}
\begin{split}
\vect{v}_{eb}^n   &= \begin{bmatrix} v_{eb,N}^n & v_{eb,E}^n & v_{eb,D}^n  \end{bmatrix}^T \\
\vect{p}_b        &= \begin{bmatrix} L_b & \lambda_b & h_b  \end{bmatrix}^T \\
\end{split}
\label{eq.vect_comp}
\end{equation}
The map of a distance in a curvilinear frame (latitude, longitude, and altitude) to a Cartesian frame (North, East, Down) is represented by the transformation matrix
\begin{equation}
\vect{T}_{r(n)}^p = \begin{bmatrix} \frac{1}{R_N+h_b} & 0 & 0 \\ 0 & \frac{1}{(R_N+h_b)\cos(L_b)} & 0 \\ 0 & 0 & -1 \end{bmatrix}
\label{eq.lla_to_ned}
\end{equation}
The rotation rate of the earth expressed in the navigation frame is
\begin{equation}\label{eq.earth_rate}
\vect{\omega}_{ie}^n = \begin{bmatrix} \omega_{ie}\cos(L_b) & 0 & -\omega_{ie}\sin(L_b) \end{bmatrix}^T
\end{equation}
where the rotation rate of the earth around its axis is $\omega_{ie} = 7.292115\e{-5}\ \si{\radian\per\second}$. The transport rate is
\begin{equation}\label{eq.transport_rate}
\vect{\omega}_{en}^n = \begin{bmatrix} \frac{v_{eb,E}^n}{R_E + h_b} & \frac{v_{eb,N}^n}{R_N + h_b} & -\frac{v_{eb,E}^n \tan(L_b)}{R_E + h_b} \end{bmatrix}^T
\end{equation}
Integration of discrete IMU measurements into estimates of vehicle attitude, velocity, and position require discrete-time approximations of \eqref{eq.nav_equations}. Detailed derivations of the discrete-time expressions are given in \cite{groves2013principles} and are not included here for brevity.

%===============================================================================
% Sensor Models
%===============================================================================
\section{Sensor Models}\label{sec.sensor_models}

\begin{figure*}[t]
\centering 
\includegraphics[width=\linewidth]{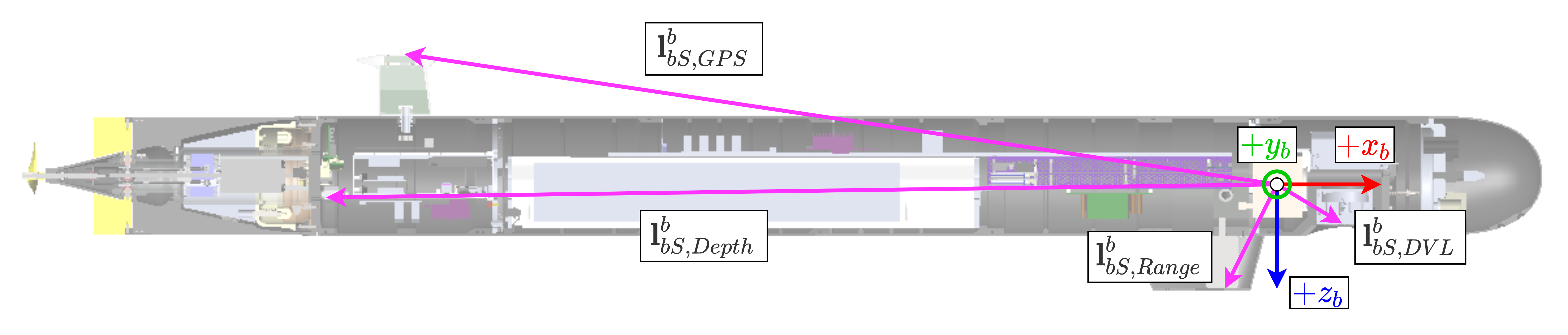}
\caption{Lever arms for the navigation sensors in the 690 AUV.}
\label{fig.690_lever_arms}
\end{figure*}

In this section, we present the sensor models used for each of the navigation sensors on the 690 AUV. Without loss of generality, the vehicle body frame is taken to be the same as the IMU sensor frame. Each additional sensor, located at different points across the AUV, has its own sensor frame denoted with an $S$ in variable subscripts and superscripts. Because the sensor frames are not co-located with the body frame, the lever arms between the body frame and the sensor frames must be accounted for. The lever arms for each sensor on the 690 AUV are shown in Figure \ref{fig.690_lever_arms}.
The corresponding value for each lever arm is given in Table \ref{tbl.690_lever_arms}.
\begin{table}[h]
\renewcommand{\arraystretch}{1.3}
\caption{Lever arms from body frame to sensor frames.}
\label{tbl.690_lever_arms}
\centering
\begin{tabular}{lccc}
\hline
                          & $x_{bS} (m)$ & $y_{bS} (m)$ & $z_{bS} (m)$ \\
$\mathbf{l}_{bS,DVL}^b$   &  0.0984      &       0      &  0.0548 \\
$\mathbf{l}_{bS,Depth}^b$ & -1.4192      & -0.0254      & -0.0156 \\
$\mathbf{l}_{bS,Range}^b$ & -0.0811      &       0      &  0.1678 \\
$\mathbf{l}_{bS,GPS}^b$   & -1.2934      &       0      & -0.1926 \\
\hline
\end{tabular}
\end{table}

These lever arms are used in each sensor's measurement model to account for the position and velocity effects of the offset sensor frame. The sensor frame velocity and position, accounting for the lever arm between the body frame and the sensor frame, are
\begin{equation}\label{eq.sensor_frame_pos}
\begin{split}
	\vect{p}_S        &= \begin{bmatrix} L_S & \lambda_S & h_S \end{bmatrix}^T = \vect{p}_b + \vect{T}_{r(n)}^p \vect{C}_b^n \vect{l}_{bS}^b \\
	\vect{v}_{eS}^{S} &= \begin{bmatrix} \vect{v}_{eS,x}^{S} & \vect{v}_{eS,y}^{S} & \vect{v}_{eS,z}^{S} \end{bmatrix}^T = \vect{v}_{eb}^{b} + \vect{\omega}_{eb}^b \wedge \vect{l}_{bS}^b
\end{split}
\end{equation}
where $\vect{l}_{bS}^b$ is the lever arm from the body frame to the sensor frame, $\vect{\omega}_{eb}^b$ is the angular velocity of the body frame relative to the earth, and $\wedge$ is the cross product. Since $\vect{\omega}_{eb}^b$ is not instrumented directly, it must be calculated from the IMU measurement of $\vect{\omega}_{ib}^b$ using
\begin{equation}
\begin{split}
	\vect{\omega}_{eS}^S &= \vect{\omega}_{eb}^b \\
	                     &= \vect{\omega}_{ib}^b - \vect{\omega}_{ie}^b \\
	                     &= \vect{\omega}_{ib}^b - \vect{C}_n^b \vect{\omega}_{ie}^n
\end{split}
\end{equation}
where $\vect{\omega}_{ie}^n$ is given by (\ref{eq.earth_rate}). 

The following subsections describe the measurement model for each navigation sensor on the Virginia Tech 690 AUV.

\subsection{Inertial Measurement Unit}\label{sec.imu_model}

Measurements of angular velocity $\tilde{\vect{\omega}}_{ib}^b$ and linear acceleration $\tilde{\vect{f}}_{ib}^b$ of the body frame relative to the inertial frame from the IMU are modeled with a constant bias and sensor noise as follows:
\begin{equation}
\begin{split}
	\tilde{\vect{\omega}}_{ib}^b &= \vect{\omega}_{ib}^b + \vect{b}_g + \vect{\eta}_g \\
	\tilde{\vect{f}}_{ib}^b &= \vect{f}_{ib}^b + \vect{b}_a + \vect{\eta}_a
\end{split}
\end{equation}
where $\vect{b}_g, \vect{b}_a \in \mathbb{R}^{3\times1}$ are gyroscope and accelerometer bias states, respectively, and 
\begin{equation}
\begin{split}
	\vect{\eta}_g \sim \mathcal{N}(\vect{0}, \sigma_g^2\eye{3}) \\
	\vect{\eta}_a \sim \mathcal{N}(\vect{0}, \sigma_a^2\eye{3})
\end{split}
\end{equation}
are gyroscope and accelerometer measurement noise. The gyroscope and accelerometer measurement noise covariance were measured to be
\begin{equation}
\begin{split}
	\sigma^2_g &= 8.4616 \e{-10} \; \si{\radian\squared\per\second\squared}\\
	\sigma^2_a &= 6.1549 \e{-5} \; \si{\meter\squared\per\second\tothe{4}}\\
\end{split}
\end{equation}
for the fiber-optic gyroscope (FOG) IMU in the 690 AUV.
\subsection{Doppler Velocity Log}

The DVL provides measurements $\tilde{\vect{v}}_{eS}^{S}$ of the linear velocity of its sensor frame $S$ relative to the earth. Assuming the DVL sensor frame axes are aligned with the body frame axes, the DVL sensor frame measurement is modeled
\begin{equation}
	\tilde{\vect{v}}_{eS}^{S} = \vect{v}_{eS}^{S} + \vect{\eta}_v
\end{equation}
The DVL sensor noise $\vect{\eta}_v$ is modeled $\vect{\eta}_v \sim \mathcal{N}(\vect{0}, \sigma^2_v \eye{3})$ with $\sigma^2_v = 0.0001\ \si{\meter\squared\per\second\squared}$.

\subsection{Depth Sensor}

A depth sensor measures the vertical distance from the surface of the water to the sensor frame of the depth sensor. To get the depth of the body frame, the lever arm from the body frame to the sensor frame of the depth sensor must be used. The depth measured by the depth sensor, $\tilde{d}$, is
\begin{equation}
\tilde{d} = h_{sea} - h_S + \eta_d
\end{equation}
where $h_{sea}$ is the altitude of the water's surface and $\eta_d$ is depth sensor noise distributed as $\eta_d \sim \mathcal{N}(0,\sigma^2_d)$ with $\sigma^2_d = 7.0\e{-5}\ \si{\meter\squared}$ for the depth sensor on the 690 AUV measured experimentally.

\subsection{GPS Receiver}

The 690 AUV is equipped with an antenna mast that includes a GPS antenna and a corresponding GPS receiver within the vehicle. The origin of the GPS frame is centered at the GPS antenna. The GPS position measurement is modeled with measurement noise as
\begin{equation}
\tilde{\vect{p}}_S = \vect{p}_S + \vect{\eta}_p
\end{equation}
where $\vect{\eta}_p \sim \mathcal{N}(\vect{0}, (\sigma^2_d[\vect{T}_{r(n)}^p]^2)$ is GPS position measurement noise with $\sigma^2_d = 6.25\ \si{\meter\squared}$ for the GPS receiver on the 690 AUV.

\subsection{Acoustic Ranging}
	
The 690 AUV and support vessel are both equipped with an acoustic modem capable of one-way travel time (OWTT) acoustic ranging. The AUV periodically receives acoustic messages from the research vessel that contain the time-of-departure of the message and the research vessel transducer's GPS location. When this message is received by the AUV, the time-of-arrival is recorded and the total time-of-flight is calculated. From the time of flight and the speed of sound in water, the range from the research vessel to the AUV can be calculated.

The acoustic range measurement is modeled as the Euclidean distance, expressed in geodetic coordinates, between the acoustic transducer sensor frame position, $\vect{p}_S$, and the research vessel transducer position embedded in the acoustic message, $\vect{p}_t$.
\begin{equation}
\tilde{r} = || \vect{T}_{r(n)}^p (\vect{p}_t - \vect{p}_S) || + \eta_r
\end{equation}
where the sensor frame position is calculated using \eqref{eq.sensor_frame_pos} and $\eta_r \sim \mathcal{N}(0, \sigma_r^2)$ with $\sigma^2_r = 1.0\ \si{\meter\squared}$.

%===============================================================================
% Unscented Kalman Filter on Manifolds
%===============================================================================
\section{Unscented Kalman Filter on Manifolds}\label{sec.manifold_ukf}

The unscented Kalman Filter on manifolds (UKF-M) is a modification to the UKF that is able to estimate the mean and covariance of a state that evolves on a Lie group \cite{brossard2020code}. This is particularly useful for estimation of vehicle attitude which is described by the Lie group $SO(3)$. In this work, the states that we wish to estimate are
\begin{equation}\label{eq.state_vector}
\begin{split}
\hat{\vect{C}}_n^b    &\in SO(3) \\
\hat{\vect{v}}_{eb}^n &\in \mathbb{R}^3 \\
\hat{\vect{p}}_b      &\in \mathbb{R}^3 \\
\hat{\vect{b}}_g      &\in \mathbb{R}^3 \\
\hat{\vect{b}}_a      &\in \mathbb{R}^3 \\
\end{split}
\end{equation}
which form the combined state $\hat{\vect{x}} \in SO(3) \times \mathbb{R}^{12}$. Implementation of the UKF-M requires a definition of a retraction $\varphi$ that maps uncertainty in the tangent space to the manifold, and its inverse $\varphi^{-1}$. Following the example presented in \cite{brossard2020code} for the mixed case where some states evolve in $SO(3)$ and others do not, we choose the retraction
\begin{equation}
	\varphi(\vect{x},\vect{\xi}) = 
	\begin{bmatrix} 
		\exp(\vect{\xi}_{1:3})\vect{C}_b^n \\ 
		\vect{v}_{eb}^n + \vect{\xi}_{4:6} \\ 
		\vect{p}_b + \vect{T}_{r(n)}^p \vect{\xi}_{7:9} \\ 
		\vect{b}_g + \vect{\xi}_{10:12}\\ 
		\vect{b}_a + \vect{\xi}_{13:15} 
	\end{bmatrix}
\end{equation}
and the corresponding inverse retraction
\begin{equation}
	\varphi^{-1}(\vect{x},\hat{\vect{x}}) = 
	\begin{bmatrix} 
		\log(\hat{\vect{C}}_b^{n} (\vect{C}_b^n)^T) \\ 
		\hat{\vect{v}}_{eb}^n - \vect{v}_{eb}^n \\ 
		\vect{T}_p^{r(n)} (\hat{\vect{p}}_b - \vect{p}_b) \\ 
		\hat{\vect{b}}_g - \vect{b}_g  \\ 
		\hat{\vect{b}}_a - \vect{b}_a 
	\end{bmatrix}
\end{equation}
where $\exp$ and $\log$ are the exponential and logarithm functions on $SO(3)$ defined in \cite{barfoot2017state} and $\vect{\xi}_{a:b}$ is a vector made of the $a$th to the $b$th element of $\vect{\xi}$. The map $\vect{T}_{r(n)}^p$ of a distance in a Cartesian frame to a curvilinear frame, the inverse of \eqref{eq.lla_to_ned}, is given by
\begin{equation}
\vect{T}_p^{r(n)} = \begin{bmatrix} R_N+h_b & 0 & 0 \\ 0 & (R_N+h_b)\cos(L_b) & 0 \\ 0 & 0 & -1 \end{bmatrix}
\label{eq.ned_to_lla}
\end{equation}
These retraction and inverse retraction functions are the based on those chosen in \cite{brossard2020code}, with two modifications. The first modification is the addition of the terms for the gyroscope and accelerometer bias states. Second, the terms in $\varphi$ and $\varphi^{-1}$ corresponding to the position state include the transformations between curvilinear and Cartesian distance. The addition of the transformations changes the position covariance estimated by the filter to be in units of meters squared with a standard deviation in meters, which is more useful for a human operator.

The prediction step of the UKF-M is driven by the arrival of IMU measurements of angular velocity and linear acceleration of the body frame relative to the inertial frame. For each measurement received from the IMU, the UKF-M prediction step is executed where the nonlinear model $f$ is the entire IMU measurement integration process from Section \ref{sec.inertial_navigation_equations}. When measurements from the other aiding sensors described in Section \ref{sec.sensor_models} are received, the UKF-M update step is executed to correct the inertial navigation state estimate. 

%===============================================================================
% Outlier Rejection
%===============================================================================
\section{Outlier Rejection}\label{sec.outlier_rejection}

\begin{figure}[h]
\centering 
\includegraphics[width=\linewidth]{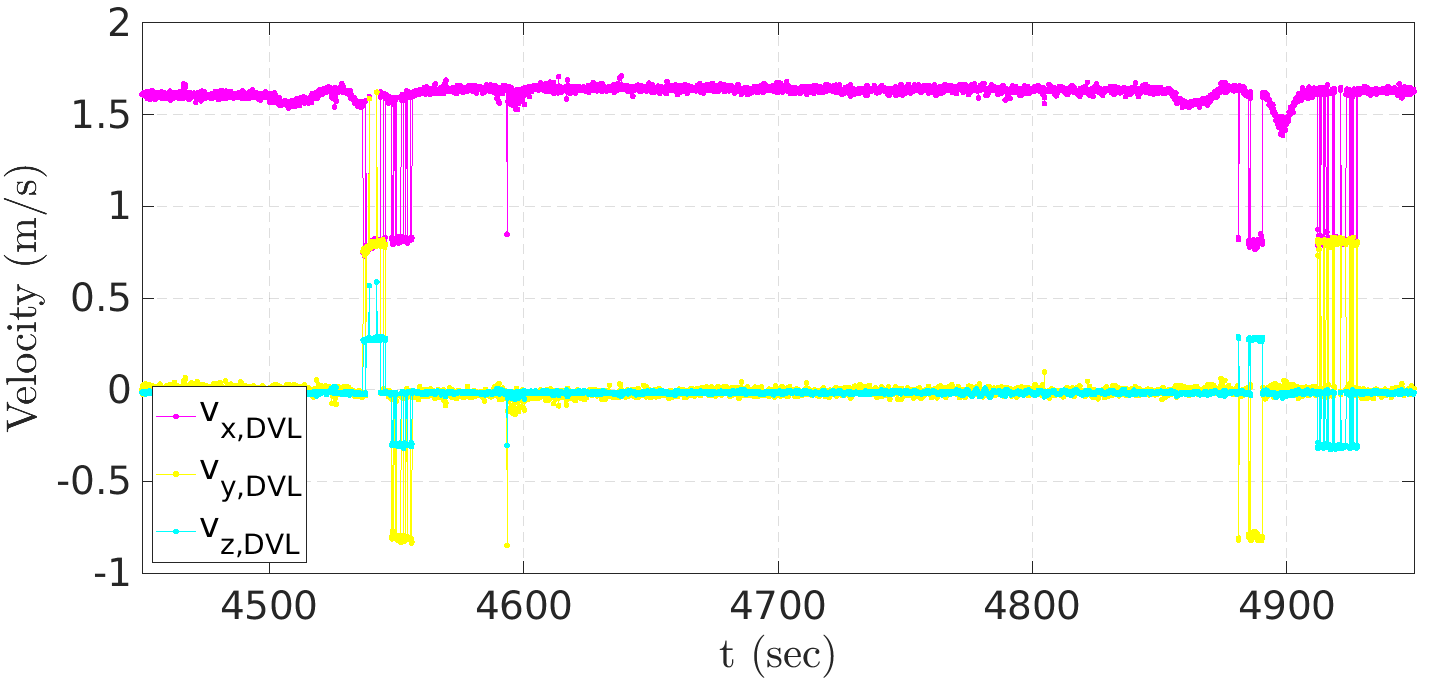}
\caption{Outliers in DVL measurements during an AUV mission.}
\label{fig.dvl_outliers}
\end{figure}

Some sensors used for AUV navigation are susceptible to erroneous measurement outliers that can significantly degrade the navigation estimate if they are used to update the filter state. Measurements involving acoustic signals are particularly prone to these outliers due to multipath and other effects that can cause measurement errors. Figure \ref{fig.dvl_outliers} shows an example of DVL measurements during a portion of an AUV mission. These outliers deviate significantly from the mean velocity values of about 1.6 m/s for the x-axis and 0 m/s for the y- and z-axis.

A simple approach to rejecting these outliers is described in \cite{krauss2020detection} and \cite{jabari2016range}, which uses the measurement innovation and a chosen threshold to decide whether the measurement should be used. A measurement with
\begin{equation}
	|\inno| > T_r
\end{equation}
where $\inno$ is the measurement innovation and $T_r$ is the threshold, is rejected. The measurement innovation is already calculated in the UKF-M update step for every measurement, requiring minimal extra computations to implement this outlier rejection. However, a threshold must be chosen high enough that legitimate measurements are used and low enough that outliers are rejected. In this work, we formulate a threshold based on the measurement covariance, $\vect{P}_{yy}$, which is also calculated during the UKF-M update step:
\begin{equation}
	T_r = M \sqrt{||\diag(\vect{P}_{yy})||}
\end{equation}
where $\diag(\vect{P}_{yy})$ is a vector of the diagonal elements of $\vect{P}_{yy}$ and $M$ is a multiplier used to increase or decrease the threshold. The tuning of $M$ presents a trade-off between sensitivity to outliers and probability of rejecting a legitimate measurement.

%===============================================================================
% Experiment Setup
%===============================================================================
\section{Experimental Results}\label{sec.experimental_results}

In order to validate the UKF-M for AUV aided inertial navigation with real data, the UKF-M presented in this work was implemented on the Virginia Tech 690 AUV's ROS-based vehicle control software framework in order to run in real time. In addition, the navigation algorithm implementation also records all navigation sensor data for post-mission analysis and post-processing. The Virginia Tech 690 AUV and research vessel were deployed in Claytor Lake, Dublin, Virginia, USA. The UKF-M state vector was initialized to all zeros except for position, which was initialized to the first GPS measurement received. 

\begin{figure}[h]
\centering 
\includegraphics[width=\linewidth]{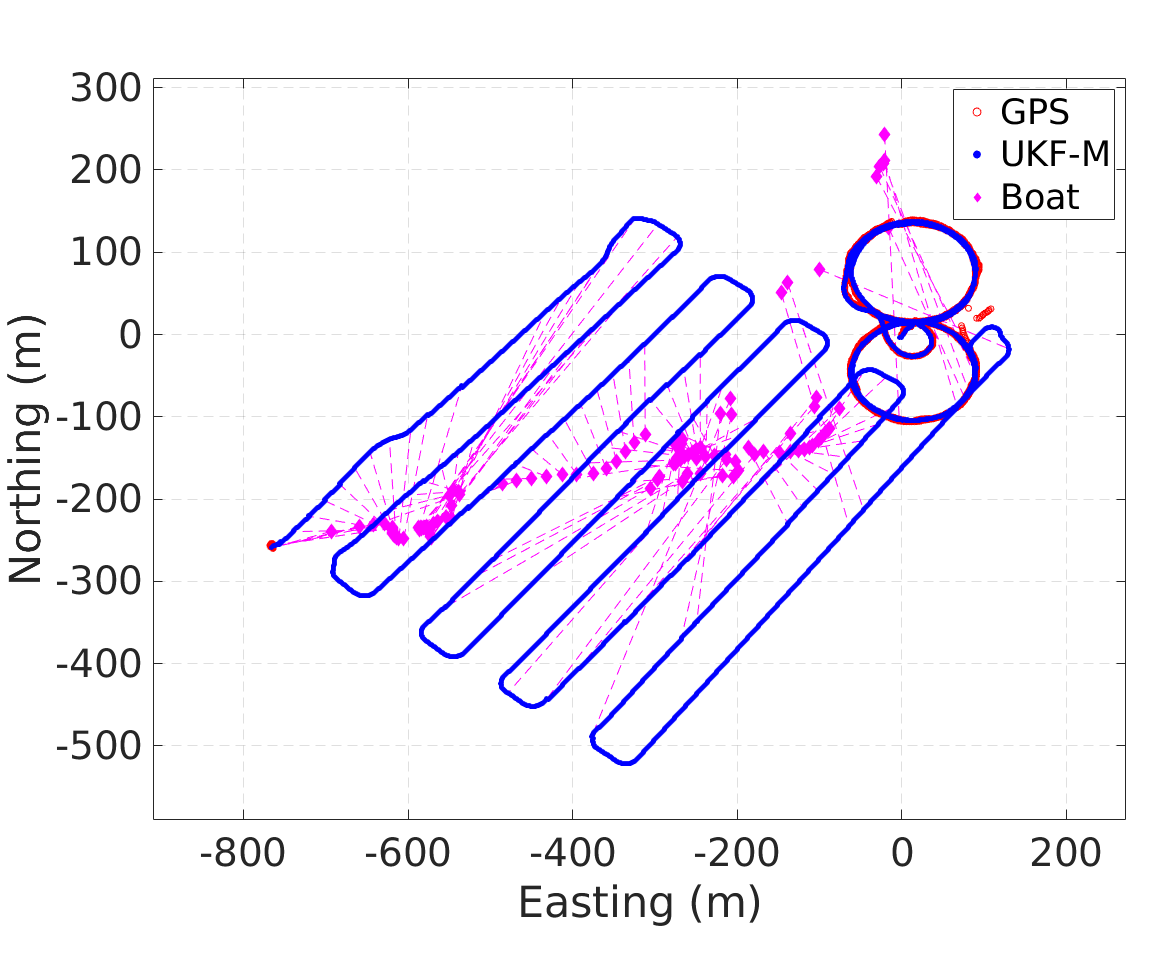}
\caption{Maneuvers executed by the 690 AUV at Claytor Lake consisting of 30 minutes of figure eight on the surface and 1 hour of lawnmower survey. Dashed lines represent an acoustic range measurement from the research vessel to the AUV.}
\label{fig.figure_eight_survey}
\end{figure}

The AUV was commanded to execute a mission consisting of a 30 minute figure-eight alignment maneuver on the surface with access to GPS followed by a survey pattern at a depth of 3 meters for approximately one hour with no GPS. The survey pattern consisted of 9 swaths of one kilometer length spaced 50 meters apart. The UKF-M estimate position, GPS measurements, and research vessel position during the AUV mission are pictured in Figure \ref{fig.figure_eight_survey} where latitude and longitude have been converted to NED position with origin latitude and longitude at the mission start position.

During the AUV mission, the research vessel drifted slowly within the survey area, broadcasting an acoustic ranging packet every 15 seconds. Approximately 45\% of these acoustic packets were received by the AUV and successfully used for an acoustic range measurement in the UKF-M measurement update step.

The error between the UKF-M position estimate and the first GPS measurement acquired upon resurfacing at the end of the mission was approximately 6.64 meters. For a distance traveled of 9400 meters, this corresponds to an error of 0.071\% of distance traveled. It should be noted that, while this error value indicates that the filter can achieve high accuracy, it is not a statistical characterization of the performance of the navigation system, which is beyond the scope of this work.

To demonstrate the ability of the UKF-M to handle large initial heading errors, the sensor data collected during the mission was post-processed multiple times through the same UKF-M with the initial heading angle set to a range of heading angles from 0 degrees to 360 degrees in 30 degree increments.

\begin{figure}[h]
\centering 
\includegraphics[width=\linewidth]{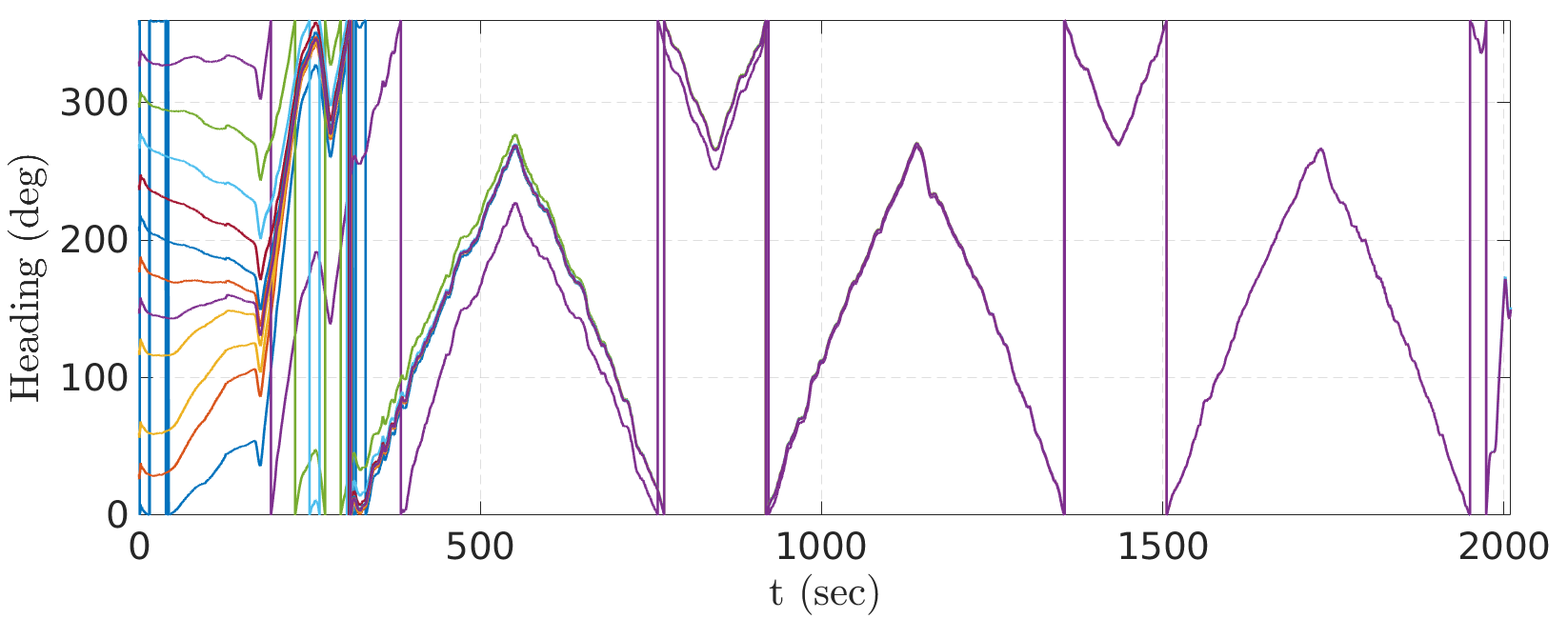}
\caption{Convergence in heading for a 360 degree range in initial heading angles.}
\label{fig.heading_convergence}
\end{figure}

Figure \ref{fig.heading_convergence} shows the UKF-M estimated heading angle over the duration of the figure-eight surface alignment maneuver for each initial heading angle case. In each case, the estimated heading converges to the same value over the course of the alignment maneuver. However, Figure \ref{fig.heading_convergence} also shows that the initial heading angle can affect the amount of alignment time required for heading to converge. 

\begin{figure}[h]
\centering 
\includegraphics[width=\linewidth]{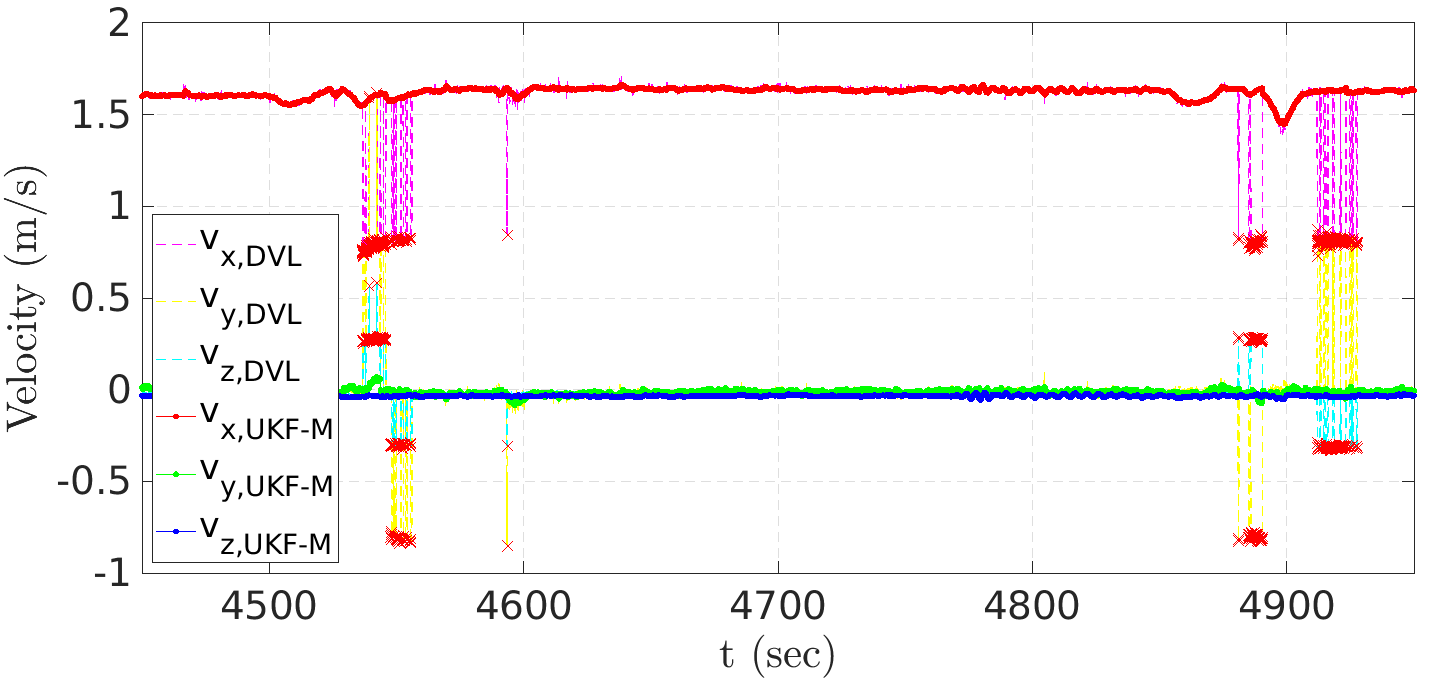}
\caption{Rejection of outliers in DVL measurements during an AUV mission. Rejected measurements are marked with a red x.}
\label{fig.dvl_outliers_rejection}
\end{figure}

In Figure \ref{fig.dvl_outliers_rejection}, the rejection of DVL measurement outliers via the technique described in Section \ref{sec.outlier_rejection} is shown. Measurements that deviate significantly from the expected velocity measurement are rejected, indicated by a red x on the measurement. As a result, the vehicle velocity estimated by the UKF-M is not affected by the outliers.

%===============================================================================
% Conclusion
%===============================================================================
\section{Conclusion}\label{sec.conclusion}

In this work, we present an implementation of the UKF-M for localization of an AUV with sensor measurement models specifically accounting for sensor lever arms and IMU bias terms. A real-time implementation of the proposed UKF-M was validated in the field using the Virginia Tech 690 AUV. The experimental results in Section \ref{sec.experimental_results} demonstrate the ability of the UKF-M to handle estimation of a combination of states that evolve in $SO(3)$ and those that evolve in $\mathbb{R}^3$. An additional measurement validation step added to the filter's measurement update step facilitates the rejection of measurement outliers that would otherwise impact the navigation state estimate.

Additionally, we experimentally demonstrate the ability of the filter heading estimate to converge despite arbitrary initial heading error. The ability to handle large initial alignment alleviates the need for separate coarse and fine alignment steps that are common when using an EKF for inertial navigation. In the experiments presented in this work, alignment is done by executing maneuvers on the surface while the UKF-M is running. Analysis of how different sets of sensors and maneuvers affect the speed and accuracy of alignment is the subject of future work.

%===============================================================================
% Bibliography
%===============================================================================
\addtolength{\textheight}{-12cm}  % This command serves to balance the column lengths

\bibliography{bib/sources} 
\bibliographystyle{ieeetr}

\end{document}